# Probabilistic Bearing Fault Diagnosis Using Gaussian Process with Tailored Feature Extraction

Mingxuan Liang

Assistant Professor

College of Mechanical and Electrical Engineering

China Jiliang University

Email: leomx@cjlu.edu.cn

Kai Zhou[†]

Research Assistant Professor

Department of Mechanical Engineering-Engineering Mechanics

Michigan Technological University

Email: kzhou@mtu.edu

---

[†] Corresponding author

# Probabilistic Bearing Fault Diagnosis Using Gaussian Process with Tailored Feature Extraction

Mingxuan Liang[a] and Kai Zhou[b+]

a: College of Mechanical and Electrical Engineering, China Jiliang University, Hangzhou 310018, China

b: Department of Mechanical Engineering-Engineering Mechanics, Michigan Technological University, Houghton, MI 49931, USA

**ABSTRACT**

Rolling bearings are subject to various faults due to its long-time operation under harsh environment, which will lead to unexpected breakdown of machinery system and cause severe accidents. Deep learning methods recently have gained growing interests and extensively applied in the data-driven bearing fault diagnosis. However, current deep learning methods perform the bearing fault diagnosis in the form of deterministic classification, which overlook the uncertainties that inevitably exist in actual practice. To tackle this issue, in this research we develop a probabilistic fault diagnosis framework that can account for the uncertainty effect in prediction, which bears practical significance. This framework fully leverages the probabilistic feature of Gaussian process classifier (GPC). To facilitate the establishment of high-fidelity GPC, the tailored feature extraction with dimensionality reduction method can be optimally determined through the cross validation-based grid search upon a prespecified method pool consisting of various kernel principal component analysis (KPCA) methods and stacked autoencoder. This strategy can ensure the complex nonlinear relations between the features and faults to be adequately characterized. Furthermore, the sensor fusion concept is adopted to enhance the diagnosis performance. As compared with the traditional deep learning methods, this proposed framework usually requires less labeled data and less effort for parameter tuning. Systematic case studies using the publicly accessible experimental rolling bearing dataset are carried out to validate this new framework. Various influencing factors on fault diagnosis performance also are thoroughly investigated.

## 1. Introduction

Rolling bearing is one type of important components in the machinery system. Its functional state directly determines the system integrity and performance, and thus needs to be strictly monitored, controlled and maintained. This generally can be realized by carrying out the bearing fault diagnosis. The timely and accurate diagnosis of rolling bearing faults recently has attracted growing interests in research community [1–3]. To facilitate the effective fault diagnosis, various types of signals can be employed. Among them, vibration signals are most commonly used as they can be easily measured from off-the-shelf sensors, and usually contain a variety of signature information that can indicate the fault occurrence [4–6]. With the advancement of computational intelligence technology, the data-driven machine learning methods have been considerably applied in fault diagnosis since they are capable of characterizing the inherent correlation between the measurement and fault conditions [7–10]. Once the machine learning models are established, the fault diagnosis can be implemented in an efficient manner. As one category of machine learning methods, deep learning models can streamline and automate the whole fault diagnosis process without any pre-processing steps required because of their excellent ability for feature extraction of raw real-time data. For this reason, the deep learning methods, such as convolutional neural network, stacked autoencoder and deep belief network have become the mainstream of bearing fault diagnosis approaches nowadays. Zhang et al integrated new training method into a deep convolutional neural network for bearing fault diagnosis under noisy environment and different working loads. This integrated model can achieve high accuracy even without any information of the target domain [11]. Liang et al developed a parallel convolutional neural network that can take full advantage of future fusion. This new methodology has been validated with better stability and robustness than other existing methods [12]. Xu et al proposed to use the stacked denoise autoencoder together with Gath-Geva clustering algorithm for bearing fault diagnosis based upon unlabeled data which are commonly seen in practical applications [13]. Shao et al developed a novel optimization framework for deep belief network training which significantly enhances the rolling bearing fault diagnosis performance [14].

Recently, different advanced signal processing techniques also have been developed, aiming at further enhancing the fault feature extraction to facilitate the deep learning analysis for bearing fault diagnosis. The most well-known category of approaches is to analyze the signals over time-frequency domain, including but not limited to fast Fourier transform (FFT) and continuous wavelet transform (CFT). Pandarakone et al established a deep learning framework based upon the features extracted from the fast Fourier transform analysis to achieve the diagnosis of practical bearing fault occurrence [15]. Xu et al developed a multi-step bearing fault diagnosis strategy using the combination of convolutional neural network and random forest ensemble, where the continuous wavelet transform was adopted for signal conversion, building the connection between different diagnosis steps [16]. While these methods have shown the promise, their effectiveness is subject to the manual feature selection. It is worth pointing out that the size of dataset plays an equal role as the features in affecting the performance of deep learning analysis. In order to capture the complex input-output relation, deep learning model oftentimes is built with large-scale, and expected to be adequately trained upon the sufficiently large-sized data. Nevertheless, the labeled data in actual practice are limited, which will lead to the model overfitting. There indeed exist some solutions to tackle the overfitting issue caused by the data scarcity. For example, some data-enhanced methods have been proposed such as overlap truncation and interpolation of raw time-series signals [11,17]. Although the data will be significantly enriched, such operation may introduce the inter-dependency of data used for model training. Data augmentation that acts as a regularizer can mitigate the overfitting by adding slightly modified copies of existing data [18,19]. In addition to approaches with specific focus on data enrichment, another different class of approaches aim to reconfigure/redesign the deep learning model architecture or learning strategy to achieve the desired performance given limited data. Transfer learning and semi-supervised learning are two most representative approaches [20–23]. The former one, i.e., transfer learning allows the deep learning model to be trained through a previous/source task, and then repurposed as the initial model for a new/destination task. The hypothesis to ensure the success of transfer leaning is that the source and destination tasks are intrinsically similar. However, such hypothesis does not always hold in most practical situations. The latter one, i.e., semi-supervised learning can fully take advantage of the large amount of unlabeled data in training and thus improve the prediction quality. Fundamentally, in this approach there exists a minimum size of labeled data required to capture the overall target distribution, which may still be relatively large in actual scenarios. It is worth mentioning that the performance of deep learning model also highly relies on its hyperparameters [24,25]. Tuning the hyperparameters essentially follows the trial-and-error procedure, which requires the extensive experience and knowledge. It usually is very time consuming when a large number of hyperparameters are subject to tuning.

The above literature review indicates that while the deep learning modes are appealing, the labeled data scarcity and hyperparameter tuning will pose significant challenges. In practical implementation, the various uncertainty sources inevitably exist, which will negatively interfere with the fault diagnosis since the fault classification is point estimation in nature [26]. In other words, the fault classification is deterministic that will likely result in the false detection/alarm. On the other hand, the probabilistic fault diagnosis can take those uncertainties into account and yield the uncertainty estimation in prediction. The rich probabilistic information in prediction allows one to further incorporate the empirical knowledge, which can lead to the informed decision. Since the probabilistic fault diagnosis tightly aligns with the decision making strategy for dealing with the real problem that are stochastic in nature, it bears practical significance. In contrast to deterministic fault diagnosis, the probabilistic fault diagnosis can provide the predictive distribution, i.e., the distribution of probability for being certain known fault condition. While in the deep learning models, such as convolutional neural network, the value from softmax layer is interpreted as probability upon which the class then will be assigned, this value does not truly mean the probability from the statistical perspective.

To enable the probabilistic fault diagnosis in this research, Bayesian approach appears to be a tailored approach. Bayesian inference has been extensively integrated into the machine learning methods as a probabilistic optimizer for model training [27]. These specific types of machine learning methods hence possess the probabilistic feature that can be potentially applied on the probabilistic fault diagnosis. One well-established method is Gaussian process (GP) which is rigorously built upon the Bayesian inference-based learning. Over the past decade, GP has been widely adopted in solving various engineering problems, most of which were formulated as regression problems [28–33]. In this research, the bearing fault diagnosis fundamentally belongs to the classification type of analysis. We therefore need to establish a Gaussian process classifier (GPC). Unlike the GP regression, the posterior in GP classification is no longer Gaussian distribution which needs additional approximation procedure [34,35]. Mathematically, training GP classifier involves a series of matrix operations, and its computational complexity hence will exponentially grow if the dataset size or input feature number becomes large. Since the vibration time-series signal used in the fault diagnosis contains numerous features, feature extraction with dimensionality reduction should be employed to alleviate the computational burden of GPC training. There are many feature extraction methods available for selection. As reported in literature, kernel principal component analysis (KPCA) is able to reveal the nonlinearity in the data, which is favorable for the machinery fault diagnosis [4,36,37]. Autoencoder as one type of unsupervised neural networks also can be used for automatic feature extraction from the data. It is a promising feature extraction tools that has been applied onto different research fields such as speech recognition, self-driving cars, human gesture recognition [38–40]. For the features with complex relationship, single autoencoder generally is insufficient. Instead, the stacked autoencoder, i.e., multiple encoders stacked on top of one another will be adopted [41]. As the fault features within the data are completely hidden, without prior knowledge it is difficult to determine which feature extraction method is most suitable for GPC to achieve the desired diagnosis performance. As such, in this research we plan to construct a feature extraction method set, in which the tailored feature extraction method can be adaptively identified via the cross-validation-based grid search while treating the feature extraction method as a tunable parameter of GPC.

The remainder of this paper is organized as follows. In Section 2, the proposed methodology is mathematically outlined. Section 3 provides the implementation details and methodology validation where the public bearing fault database from Case Western Reserve University (CWRU) bearing data center (http://csegroups.case.edu/bearingdatacenter) is utilized. The systematic performance investigation is also carried out. Concluding remarks are summarized in Section 4.

## 2. Methodology

In this section, we first succinctly present the theory behind the Gaussian process classifier (GPC) which is the mainstay of the probabilistic fault diagnosis framework. The potential methods for feature extraction with dimensionality reduction including KPCA and stacked autoencoder that will be integrated with GPC are introduced subsequently.

### *2.1. Probabilistic classification using Gaussian process*

As will be shown later, we will use a set of binary GP classifiers to approximate the multi-class GP classifier for diagnosis of various bearing fault conditions in this research since this treatment mathematically is tractable and convenient. Hence, here we will introduce the basis idea behind the binary GP classification [35]. We first let a random variable be expressed by the GP,

$$F(\mathbf{x}) \sim N(m(\mathbf{x}), k(\mathbf{x},\mathbf{x})) \tag{1}$$

where $\mathbf{x}$ is the input features, and $m(\mathbf{x})$ and $k(\mathbf{x},\mathbf{x})$ denote the mean and covariance functions, respectively. Those functions are critical to the underlying behavior of GPC, and thus the care should be taken on the function selection. Currently, a number of mean and covariance functions are available for modeling GPC. Mean functions include zero, constant, polynomial functions and so on. Covariance functions have more variety of choices and play a more significant role than mean functions. For demonstration, several covariance functions are as follows,

$$k(\mathbf{x}_i,\mathbf{x}_j) = \sigma_f^2 \exp[-\frac{1}{2}\frac{(\mathbf{x}_i - \mathbf{x}_j)^T(\mathbf{x}_i - \mathbf{x}_j)}{\sigma_l^2}] \tag{2a}$$

$$k(\mathbf{x}_i,\mathbf{x}_j) = \sigma_f^2 \exp[1+\frac{(\mathbf{x}_i - \mathbf{x}_j)^T(\mathbf{x}_i - \mathbf{x}_j)}{2\alpha\sigma_l^2}]^{-\alpha} \tag{2b}$$

$$k(\mathbf{x}_i,\mathbf{x}_j) = \sigma_f^2 (1+\frac{\sqrt{3(\mathbf{x}_i - \mathbf{x}_j)^T(\mathbf{x}_i - \mathbf{x}_j)}}{\sigma_l})\exp[-\frac{\sqrt{3(\mathbf{x}_i - \mathbf{x}_j)^T(\mathbf{x}_i - \mathbf{x}_j)}}{\sigma_l}] \tag{2c}$$

Equation (2a), (2b) and (2c) represent the squared exponential, rational quadratic and matern 3/2 covariance functions respectively. The parameters, i.e., $[\sigma_f, \sigma_l]$ in those functions are defined as hyperparameters which will be optimized through model training. $\sigma_f$ and $\sigma_l$ are defined as signal standard deviation and scale length, respectively. It is noted that above covariance functions also have some variants. For example, automatic relevance determination (ARD) can be integrated into each of above covariance functions, where a separate length scale is assigned for each predictor/input feature.

For classification analysis in this research, we specifically are interested in the probabilistic prediction of the fault condition/class, we can map $F(\mathbf{x})$ into the unit function with the span [0,1] and obtain

$$p_{y|\mathbf{x}}(1|\mathbf{x}) = \gamma(F(\mathbf{x})) \tag{3}$$

where $\gamma(.)$ denotes the unit function. Since $F(\mathbf{x})$ is a random variable, Equation (3) is stochastic. Assume we can covert the stochastic equation to the deterministic one in the form of

$$p_{y|F(\mathbf{x})}(1|f(\mathbf{x})) = \gamma(f(\mathbf{x})) \tag{4}$$

where $f(\mathbf{x})$ becomes a deterministic latent function. The generic form of Equation (4) that takes both classes $y=1$ and $y=0$ can be formulated as

$$p_{y|F(\mathbf{x})}(y|f_y(\mathbf{x})) = \gamma(f_y(\mathbf{x})) \tag{5}$$

There are two common choices to model the unit function given as

$$\pi(f_y(\mathbf{x})) = \frac{1}{1+\exp(-f_y(\mathbf{x}))} \tag{6a}$$

$$\Phi(f_y(\mathbf{x})) = \frac{1}{\sqrt{2\pi}} \int_{-\infty}^{f_y(\mathbf{x})} \exp(-\frac{1}{2}t^2)dt \tag{6b}$$

where $\pi(\mathbf{x})$ and $\Phi(\mathbf{x})$ are the logistic likelihood and cumulative Gaussian likelihood functions, respectively that are used to assess the probability of class. Essentially, such standard likelihood function can be considered as GP prior.

Given the training inputs $\mathbf{X} = [\mathbf{x}_1, \mathbf{x}_2, ... \mathbf{x}_P]^T$ and corresponding outputs $\mathbf{y} = [y_1, y_2, ... y_P]^T$ in the dataset, we can describe the Equation (1) with the expanded form given as

$$\mathbf{F}(\mathbf{X}) \sim N_P\left( \begin{pmatrix} m(\mathbf{x}_1) \\ m(\mathbf{x}_2) \\ ... \\ m(\mathbf{x}_P) \end{pmatrix}, \begin{bmatrix} k(\mathbf{x}_1,\mathbf{x}_1) & k(\mathbf{x}_1,\mathbf{x}_2) & ... & k(\mathbf{x}_1,\mathbf{x}_P) \\ k(\mathbf{x}_2,\mathbf{x}_1) & .. & ... & k(\mathbf{x}_2,\mathbf{x}_P) \\ ... & ... & .. & ... \\ k(\mathbf{x}_P,\mathbf{x}_1) & k(\mathbf{x}_P,\mathbf{x}_2) & ... & k(\mathbf{x}_P,\mathbf{x}_P) \end{bmatrix} \right) \tag{7}$$

where $P$ is the number of data points. Assume the outputs are independent when conditioning on $\mathbf{F}(\mathbf{X})$, from which we can obtain the conditional probability of entire training dataset,

$$p_{\mathbf{y}|F(\mathbf{X})}(\mathbf{y}|f(\mathbf{X})) = \prod_{i=1}^{P} p_{y_i|F(\mathbf{x}_i)}(y_i|f(\mathbf{x}_i)) \tag{8}$$

For the notation brevity, we let $F_i$ and $f_i$ denote $F(\mathbf{x}_i)$ and $f(\mathbf{x}_i)$, respectively. Also, we let $\mathbf{F}$ and $\mathbf{f}$ denote $F(\mathbf{X})$ and $f(\mathbf{X})$, respectively. Equation (7) then can be simplified as

$$p_{\mathbf{y}|\mathbf{F}}(\mathbf{y}|\mathbf{f}) = \prod_{i=1}^{P} p_{y_i|F_i}(y_i|f_i) \tag{9}$$

Suppose we have the new testing input denoted as $\mathbf{x}^*$, we can compute the probability density function (PDF) of the conditional distribution of $\mathbf{F}^*$ by mean of marginalization,

$$p_{F^*|\mathbf{y}}(f^* | \mathbf{y}) = \int_{\mathbb{R}^P} p_{F^*|\mathbf{y}}(f^*, \mathbf{f} | \mathbf{y}) d\mathbf{f} \tag{10}$$

Using Bayes' theorem, we can obtain $p_{F^*|\mathbf{y}}(f^*, \mathbf{f} | \mathbf{y}) = \dfrac{p_{\mathbf{y}|F^*,\mathbf{F}}(\mathbf{y} | f^*, \mathbf{f}) p_{F^*,\mathbf{F}}(f^*, \mathbf{f})}{p_{\mathbf{y}}(\mathbf{y})}$. Substituting it into Equation (10) yields

$$p_{F^*|\mathbf{y}}(f^* | \mathbf{y}) = \int_{\mathbb{R}^P} p_{F^*|\mathbf{F}}(f^* | \mathbf{f}) p_{\mathbf{F}|\mathbf{y}}(\mathbf{f} | \mathbf{y}) d\mathbf{f} \tag{11}$$

where $p_{\mathbf{F}|\mathbf{y}}(\mathbf{f} | \mathbf{y})$ is the posterior over the latent function. It is worth mentioning that the analytical solution of Equation (11) does not exist because $p_{\mathbf{F}|\mathbf{y}}(\mathbf{f} | \mathbf{y})$ is a non-Gaussian distribution. The approximation technique thus is required to identify $p_{\mathbf{F}|\mathbf{y}}(\mathbf{f} | \mathbf{y})$, which also is referred to as Gaussian approximation. Its general idea is to optimally identify a gaussian distribution $q_{\mathbf{F}|\mathbf{y}}(\mathbf{f} | \mathbf{y})$ to replace $p_{\mathbf{F}|\mathbf{y}}(\mathbf{f} | \mathbf{y})$. This can be realized by some well-developed approximate inference methods, such as Laplace approximation, Expectation propagation (EP), Kullback Leibler (KL)-divergence minimization and variational bounds [42]. Once we have the approximated $p_{F^*|\mathbf{y}}(f^* | \mathbf{y})$, we can obtain the quantity of interest (QOI) in this research, so called *predictive distribution* of class probability, which is defined as

$$p_{y^*|\mathbf{y}}(y^* | \mathbf{y}) = \int_{\mathbb{R}^P} p_{y^*|F^*}(y^* | f^*) p_{F^*|\mathbf{y}}(f^* | \mathbf{y}) d\mathbf{f} \tag{12}$$

where $p_{y^*|F^*}(y^* | f^*)$ is the standard likelihood function defined in Equation (4). After the general form of predictive distribution is established by using abovementioned inference method, the model training will be implemented. Recall Equation (7) that the random vector **F** depends on the hyperparameters of the specified mean and covariance functions. As a result, the predictive distribution also will be implicitly related to those hyperparameters. The purpose of the model training hence is to optimize the hyperparameters and find the best predictive prediction $\hat{p}_{y^*|\mathbf{y}}(y^* | \mathbf{y})$, upon which the fault condition can be identified/classified accordingly.

### *2.2. Data pre-processing and feature extraction using KPCA and stacked autoencoder*

The feature extraction with dimensionality reduction is a necessary pre-processing step to facilitate the subsequent GPC construction. Since the bearing fault data generally exhibits high nonlinearity with respect to fault scenarios, the feature extraction method is expected to possess the powerful capability to handle the nonlinear data. Some tailored nonlinear methods may include kernel principal component analysis (KPCA) and stacked autoencoder.

*2.2.1. KPCA*

The basic idea of KPCA is to firstly convert the input space into a feature space via nonlinear mapping, and then calculate principal components (PCs) directly in the feature space. Compared with other nonlinear methods, the main merit of KPCA is the high computational efficiency because of no optimization required. Instead, it only requires the implementation of the linear algebra. Additionally, due to its ability to utilize different kernels, it can deal with a wide range of nonlinearity. Owing to these merits, KPCA has been widely used as feature extraction method for supervised learning analyses [4,37,43].

Given a set of data $\mathbf{x}_r \in \mathbb{R}^M$, $r = 1, 2, ,, , N$, i.e., vibration time-series signals in this research, the nonlinear correlations can be decoupled through diagonalizing their covariance matrix, which can be expressed in the reduced-dimensional linear feature space, i.e., [44]

$$\mathbf{C} = \frac{1}{P} \sum_{i=1}^{P} \phi(\mathbf{x}_i) \phi(\mathbf{x}_i)^T \quad (13)$$

where $\phi(.)$ is a nonlinear mapping function, converting the vibration signals from original input space to feature space with dimensionality reduction. In Equation (13), it is assumed that the projected new features have zero mean, i.e., $\sum_{i=1}^{P} \phi(\mathbf{x}_i) = 0$. To calculate covariance matrix $\mathbf{C}$, one may have to solve the eigenvalue problem below,

$$\mathbf{C}\mathbf{v} = \lambda \mathbf{v} \quad (14)$$

where $\lambda$ is the eigenvalue and $\mathbf{v}$ is the associated eigenvector. Substituting Equation (14) into (13), yields

$$\frac{1}{P} \sum_{i=1}^{P} \phi(\mathbf{x}_i) \{ \phi(\mathbf{x}_i)^T \mathbf{v} \} = \lambda \mathbf{v} \quad (15)$$

Let a new coefficient be defined as $\alpha_i = \dfrac{\phi(\mathbf{x}_i)^T \mathbf{v}}{P \lambda}$. We then can rewrite Equation (15) as

$$\mathbf{v} = \sum_{i=1}^{P} \alpha_i \phi(\mathbf{x}_i) \quad (16)$$

Combining Equation (15) and (16), we obtain

$$\frac{1}{P} \sum_{i=1}^{P} \phi(\mathbf{x}_i) \phi(\mathbf{x}_i)^T \sum_{j=1}^{P} \alpha_j \phi(\mathbf{x}_j) = \lambda \sum_{i=1}^{P} \alpha_i \phi(\mathbf{x}_i) \quad (17)$$

We can define a kernel function with dot product to represent the mapping function operation in the feature space as,

$$\kappa(\mathbf{x}_i, \mathbf{x}_j) = \phi(\mathbf{x}_i)^T \phi(\mathbf{x}_j) \quad (18)$$

There exist a number of available kernel functions. The most common kernel functions are as follows.

*Radial basis kernel* (also known as *Gaussian kernel*):

$$\kappa(\mathbf{x}_i, \mathbf{x}_j) = \exp(-\frac{\|\mathbf{x}_i - \mathbf{x}_j\|^2}{2\sigma^2}) \qquad (19)$$

where $\|.\|$ indicates the Euclidean distance of two vectors.

*Polynomial kernel*:

$$\kappa(\mathbf{x}_i, \mathbf{x}_j) = (\mathbf{x}_j^T \mathbf{x}_i + a)^b \qquad (20)$$

*Sigmoid kernel*:

$$\kappa(\mathbf{x}_i, \mathbf{x}_j) = \tanh(a\mathbf{x}_j^T \mathbf{x}_i^i + b) \qquad (21)$$

where $\sigma$, $a$, and $b$ are the kernel parameters that can be prespecified by the user. It is worth mentioning that, if the kernel is a linear function, the KPCA will degrade to the standard PCA. In other words, standard PCA is a special case of KPCA. Multiplying $\phi(\mathbf{x}_z)^T$ at both sides of Equation (17) and applying the Equation (18) yields

$$\frac{1}{P}\sum_{i=1}^{P}\kappa(\mathbf{x}_z, \mathbf{x}_i)\sum_{j=1}^{P}\alpha_j\kappa(\mathbf{x}_i, \mathbf{x}_j) = \lambda\sum_{i=1}^{P}\alpha_i\kappa(\mathbf{x}_z, \mathbf{x}_i) \qquad (22)$$

Above equation can be expressed in the matrix notation, i.e.,

$$\mathbf{K}\boldsymbol{\alpha} = P\lambda\boldsymbol{\alpha} \qquad (23)$$

where $\mathbf{K}$ is the kernel matrix. Clearly, this again can be solved through the eigenvalue analysis. Eigenvalues then can be sorted in a descend order, in which the largest and smallest eigenvalues will contribute to the first and last PCs, respectively. For $k$-th eigenvalue $\lambda_k$, we have an associated $P$-dimensional eigenvector $\boldsymbol{\alpha}_k = [a_k^1, a_k^2, ..., a_k^P]^T$. The $k$-th PC of a test vector $\mathbf{x}^*$ hence is obtained as

$$y_k(\mathbf{x}^*) = \sum_{r=1}^{P}\alpha_k^r\kappa(\mathbf{x}^*, \mathbf{x}_r) \qquad (24)$$

In actual practice, the projected features may have zero mean. Therefore, the mean centering should be performed by substituting kernel matrix $\mathbf{K}$ with

$$\mathbf{K}' = \mathbf{K} - \mathbf{1}_P\mathbf{K} - \mathbf{K}\mathbf{1}_P + \mathbf{1}_P\mathbf{K}\mathbf{1}_P \qquad (25)$$

where $\mathbf{1}_P$ denotes a $P \times P$ matrix with all elements equal to $1/P$. The new matrix $\mathbf{K}'$ is defined as gram matrix, which will replace $\mathbf{K}$ in Equation (23) for PC calculation.

### 2.2.2. Stacked autoencoder (SAE)

Autoencoder (AE) is one specific neural network designed to learn the efficient coding of the unlabeled data, which belongs to the unsupervised learning method. An autoencoder consists of two major parts, an encoder and a decoder. The purpose of the encoder is to transform the original data into the feature representation, from which the decoder is able to reconstruct the original data.

The encoder is a function $f$ that maps an input $\mathbf{x} \in \mathbb{R}^M$ to the hidden representation $f(\mathbf{x}) \in \mathbb{R}^L$ (usually $L \leq M$), The mapping relation can be mathematically described as [45]

$$f(\mathbf{x}) = a_f(\mathbf{W}_f \mathbf{x} + \mathbf{b}_f) \tag{26}$$

where $a_f(.)$ is the activation function of encoder. When the activation function is nonlinear, the autoencoder can perform the nonlinear feature transformation. $\mathbf{W}_f$ is a weight matrix with size $L \times M$ and $\mathbf{b}_f$ is the bias vector with size $L \times 1$. In comparison, the decoder is a function $g$ that maps the hidden representation $f(\mathbf{x}) \in \mathbb{R}^L$ to the output $g(f(\mathbf{x})) \in \mathbb{R}^M$ that has the same size with the input $\mathbf{x}$.

$$g[f(\mathbf{x})] = a_g[\mathbf{W}_g f(\mathbf{x}) + \mathbf{b}_g] \tag{27}$$

where $a_g(.)$ is the activation function of decoder. Unlike $a_f(.)$, $a_g(.)$ typically is a linear activation function. $\mathbf{W}_g$ is a weight matrix with size $M \times L$ and $\mathbf{b}_f$ is the bias vector with size $M \times 1$. The underlying idea of the autoencoder shown above enables its wide application in feature extraction and data denoising [41,46]. The training of autoencoder aims at identifying the weights and biases in the model that minimizes the reconstruction error on given training samples $\mathbf{X} = [\mathbf{x}_1, \mathbf{x}_2, ....\mathbf{x}_P]$ by solving the optimization problem below.

$$\arg\min\{E[L(\mathbf{X}, g(f(\mathbf{X})))]\} \tag{28}$$

where $L(.)$ denotes the reconstruction error/loss function. $E(.)$ is the expectation of errors for all training samples that is defined as cost function. In this research, we specifically want to leverage its capability of feature extraction and dimensionality reduction. The resulting features can be extracted directly from the encoding function, i.e., $f(\mathbf{x})$ after the autoencoder is well trained.

Stacked autoencoders (SAE), as the name indicates is a sequential aggregation of several autoencoders. The first autoencoder will be trained in the first place. The first encoding function of the trained first autoencoder will then be used as the input to train the second autoencoder to learn the second encoding function. The procedure will be repeated until all involved autoencoders are trained. The resulting feature representation is obtained from the last encoding function. It is worth noting that the stacked autoencoder also can be used for supervised learning purpose. In addition to the independent training of the autoencoders mentioned above, the actual output layer will be added to the network and the fine-tuning of the whole network is required to enable the supervised learning analysis.

## 3. Probabilistic Fault Diagnosis Practice on CRWU Bearing Fault Data

In this section, the proposed probabilistic fault diagnosis of rolling bearing will be practiced on the database from the Case Western Reserve University (CWRU) bearing data center. Our primary focus is to illustrate the underlying strength of the proposed methodology, i.e., probabilistic fault diagnosis. To further demonstrate its feasibility, its general classification performance will also be systematically investigated by comparing it with other state-of-the art fault diagnosis techniques.

### *3.1. Experimental data acquisition and fault labeling*

The public data provided by CWRU are acquired from the experimental test rig schematically shown in Figure 1. This test rig is comprised of an induction motor, a torque transducer, a dynamometer, and several accelerometers. The tested bearing is deep groove ball bearing (6205-2RS JEM SKF) that is installed at the drive end of the induction motor.

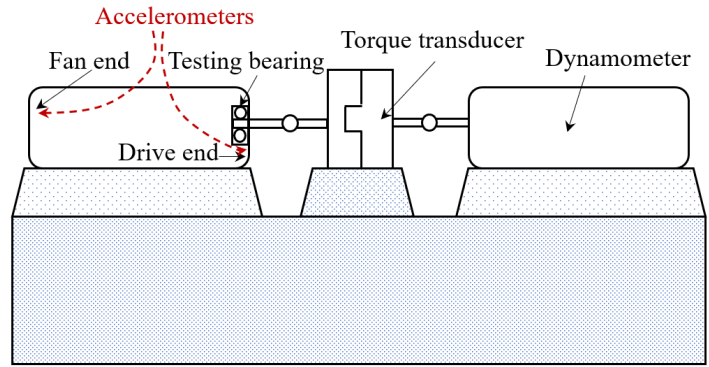

Figure 1. Experimental test rig.

Various datasets will be collected when varying the motor speed/ load and sampling frequency. In this research, we only use one dataset corresponding to 1,797 rpm motor speed and 12K sampling frequency since this already can support the methodology validation. This dataset contains 9 subsets pointing to 3 different severities, i.e., 0.007, 0.014 and 0.021 inches on the bearing fault of inner ring (IR), ball element (BA) and outer ring (OR). In addition to one healthy condition, totally 10 fault conditions are involved in the subsequent analysis. The time-domain vibration signals were recorded by the accelerometers placed at the drive end (DE) and fan end (FE) of the induction motor with magnetic bases. The bearing fault data used in this research are summarized in Table 1. The categorical labels/classes from 1 to 10 are defined for associated fault conditions. In each condition, the first 10-second vibration signal from each accelerometer is segmented into 75 samples. Each sample thus has 1,600 data points corresponding to approximately 4 revolutions. This can ensue the complete periodic vibration feature of rotating component to be adequately captured. In this research, we intend to use the combined signals from both accelerometers (DE+FE) in order to enrich the signature information with respect to the faults. It is noted that above treatment will double the number of data points in each sample, i.e., 3,200 ($1,600 \times 2$) instead of increasing the size of the dataset. With such setup, we have a total of 750 ($75 \times 10$) samples to be analyzed, aiming at identifying the fault conditions via differentiating the pivot features in data samples.

**Table 1**. Bearing fault data overfiew.

| Fault Type | Fault Severity/ Inch | Dataset Size | Defined Class |
| --- | --- | --- | --- |
| Healthy | No | 75 | 1 |
| Outer Ring (IR) Fault | 0.007 | 75 | 2 |
| | 0.014 | 75 | 3 |
| | 0.021 | 75 | 4 |
| Ball Element (BA) Fault | 0.007 | 75 | 5 |
| | 0.014 | 75 | 6 |
| | 0.021 | 75 | 7 |
| Outer Ring (OR) Fault | 0.007 | 75 | 8 |
| | 0.014 | 75 | 9 |
| | 0.021 | 75 | 10 |

### *3.2. Gaussian process model construction for probabilistic fault diagnosis analysis*

The bearing fault diagnosis essentially belongs to the type of pattern recognition tasks, which generally will be realized through the classification analysis. To align with the nature of the problem, a multi-class Gaussian process classifier (GPC) needs to be established. Because directly performing the multi-class classification analysis is hinged upon the theory behind the GPC, in this research we create a set of binary GPCs to collectively play the role of a multi-class GPC, which is so called binarization technique [34]. Two common such techniques are known as one-against-all and one-against-one, respectively [34]. In this research, we adopt the one-against-all binarization technique, where 10 binary GPCs will be constructed to identify 10 fault conditions respectively. Since GPC fundamentally is built upon the Bayesian theory, it can output the predictive distributions of probability with respect to known fault classes given any fault scenario. For implementation convenience, the distribution mean of probability will only be used to direct the final decision making. Despite the significant merit of GPC, its strong limitation is the lack of scalability. In other words, it will be subject to the curse of dimensionality, exhibiting the high computational complexity when the size or feature dimension of the dataset is large. Therefore, feature extraction and dimensionality reduction procedures need to be carried out on the data samples with long time series. Without prior knowledge, it is difficult to determine the best feature extraction method that can elucidate the causative relationship between the features and respective faults. To facilitate the high-fidelity GPC modeling, we create a feature extraction method set consisting of various KPCA methods and stacked autoencoder, where the method selection is considered as a tunable parameter. This indeed provides the high level of modeling flexibility. The implementation flow of the proposed bearing fault diagnosis can be seen in Figure 2.

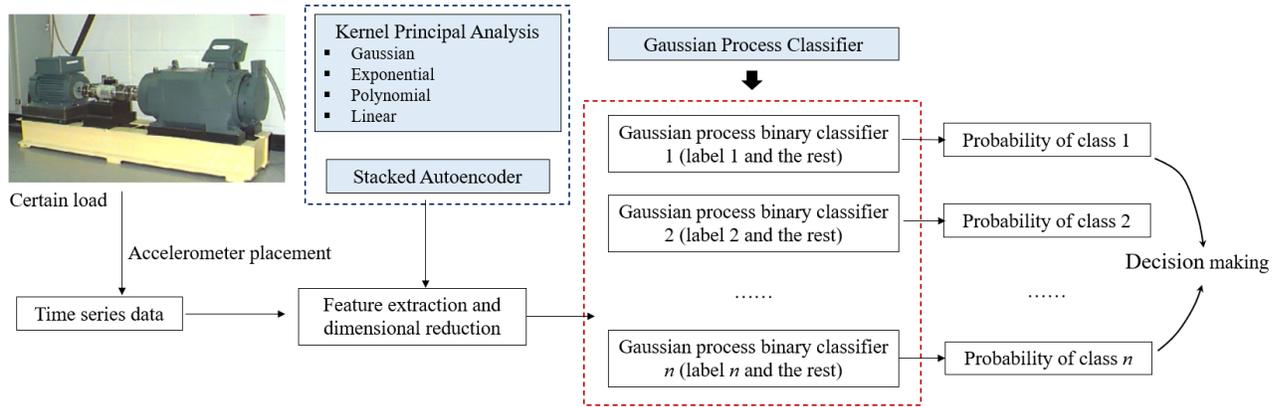

Figure 2. Implemantation flow of the bearing fault diagnosis.

In this research, the KPCA analysis transforms the data from original high-dimensional space to reduced-dimensional space using different kernel functions, including Gaussian, polynomial, linear and exponential functions. Recall that each original sample contains the signals measured by accelerometers at the drive end (DE) and fan end (FE) simultaneously. The KPCA analysis will be performed twice to obtain the associated PC information, which then will be combined together for meta-model training. This idea essentially stems from the sensor fusion concept, particularly feature-level fusion [47]. While the dimension of the transformed data sample is significantly reduced, the new data sample still is multidimensional and thus cannot be visualized completely. The lower-order principal components (PCs) usually account for the more variance of the data, which are more capable of differentiating the faults. For this reason, we here only present the first-order PCs of different kernel functions for the data measured at DE and PE as shown in Figure 3. Clearly, the distribution pattern differs with respect to the kernel function. The overlay of the PC information observed indicates that different fault conditions/classes cannot be separated regardless of kernel function. Autoencoder, as an unsupervised learning neural network model can learn the compressed representation of the raw data with massive features, and hence can be repurposed to achieve the feature extraction and dimensionality reduction purpose if configured appropriately. In this research, we specifically design a stacked autoencoder consisting of two inter-related autoencoders as shown in Figure 4a. The input of the first autoencoder represents the time-series sample, and the output has the same size of input. One hidden layer is embedded, which usually has smaller size than the input. The training of the first autoencoder is to minimize the difference, i.e., reconstruction error (Equation (28)) between the target (input) and actual output. Once trained, the feature **r** retrieved from the hidden layer of this autoencoder will be used as the input of the second one, and the same procedures will be repeated. Then the stacked autoencoder, which includes the trained hidden layers is built. The feature **q** from the hidden layer 2 will be the final feature set to be fed into the GPC. The first feature variables of both DE and FE data extracted via the stacked autoencoder are given in Figure 4b. Consistent with above observation, the boundaries of different fault classes are very unclear. This indeed poses a challenge in pursuing the desired performance of subsequent classification analysis.

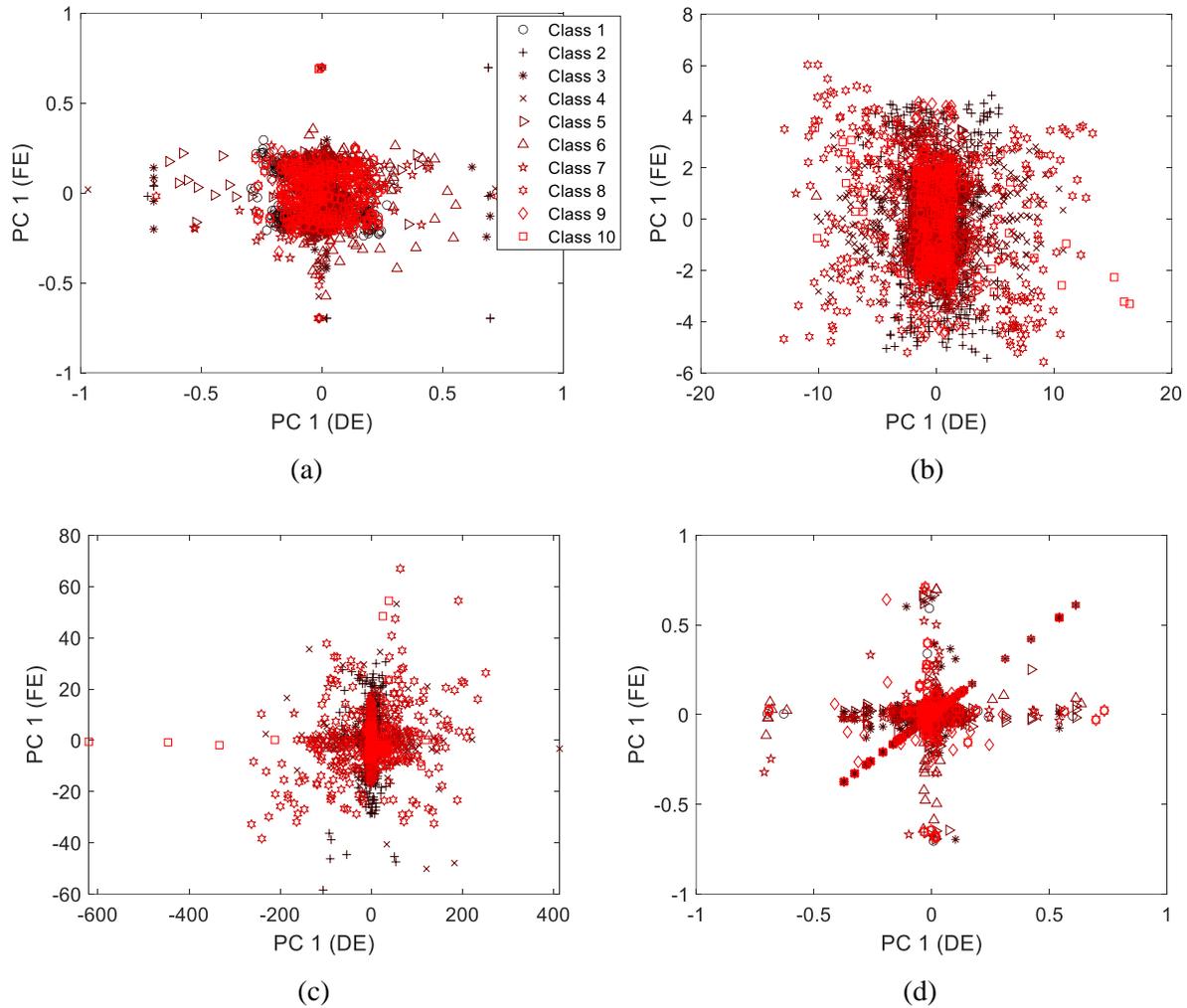

Figure 3. Illustration of first-order PCs of different fault conditions using different KPCA methods (a) Gaussian kernel; (b) Linear kernel; (c) Polynomial kernel; (d) Exponential kernel.

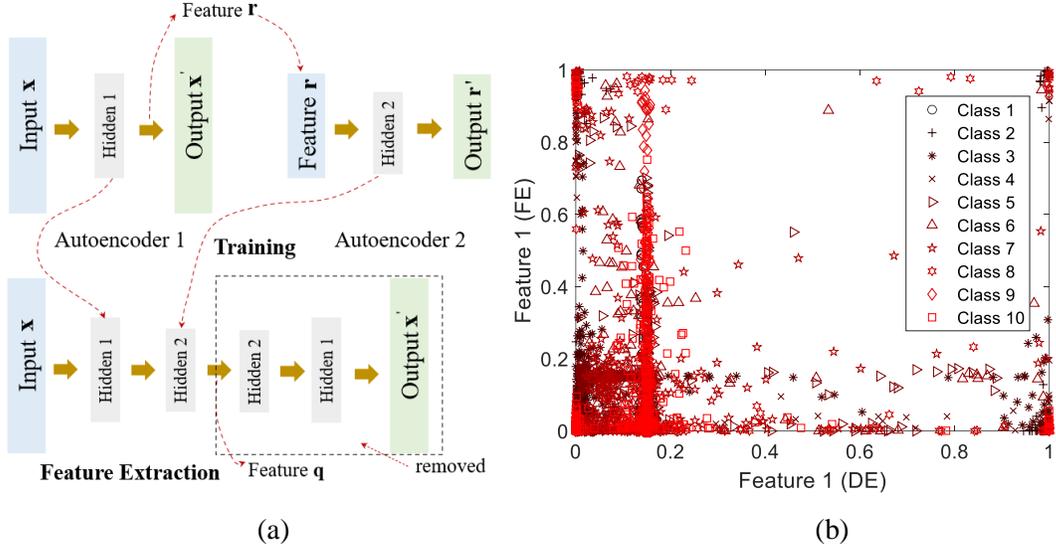

(a)                               (b)

Figure 4. Feature extraction using straked autoencoder (a) Stacked autoencoder architecture; (b) Illustration of extracted feature variables.

The key to leverage the inference capability of GPC is to incorporate the most optimal feature extraction method. Treating the feature extraction method as a tunable parameter, the method selection can be adaptively realized through the grid search in light of the GPC cross-validation performance. The cross-validation result will be detailed in subsequent subsection for self-contained presentation. The Gaussian kernel PCA is identified as the most suitable feature extraction method for GPC construction. Besides, the constant mean and ARD matern 3/2 covariance functions are used to model GPC, where KL-divergence minimization is adopted to infer the posterior. 70 out of entire 750 data samples and the rest are used for testing and training, respectively. Both training and testing subsets are ensured with class balance. After the GPC is well trained, it then will be utilized for testing/validation. The direct output of GPC in testing process is the posterior mean of probability with respect to the known fault classes. For notation convenience, we simply use "probability" to denote the posterior mean of probability throughout the manuscript. The distribution of the probability of all testing samples with respect to the fault classes is shown in Figure 5. Overall, the testing samples will have higher probability of being actual fault class than other classes. Moreover, some samples locate at the middle of the probability band. This implies that the GPC lacks confidence for these particular predictions. We then look into the details of above result i.e., the probability of testing samples from each fault class (Figure 6). Once again, it is clearly found that the samples more likely belong to their actual fault class than other classes. GPC seems to be very confident in differentiating the samples from classes 2 6 and 10. Conversely, the confidence level of classification for the samples from classes 3, 5 and 9 is relatively low. For example, the features of some samples from class 5 may also resemble the general fault signature of class 7. Such probabilistic result essentially accounts for the uncertainty in prediction, which agrees with the general logic of decision making for a stochastic process. GPC proposed in this research features the probabilistic classification, which can provide more useful information than the conventional deterministic/crispy classification. Upon the probabilistic classification outcome, the knowledge and experience can be further incorporated to interpret the uncertainty estimation, thereby assisting the final decision making.

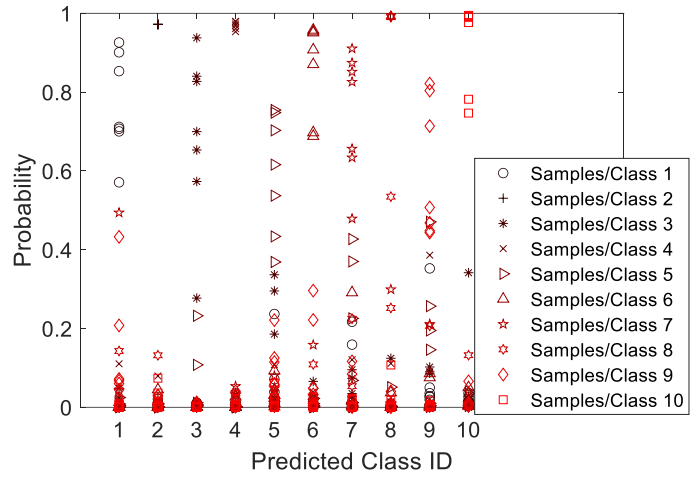

Figure 5. Probability of predicted classes over entire testing samples.

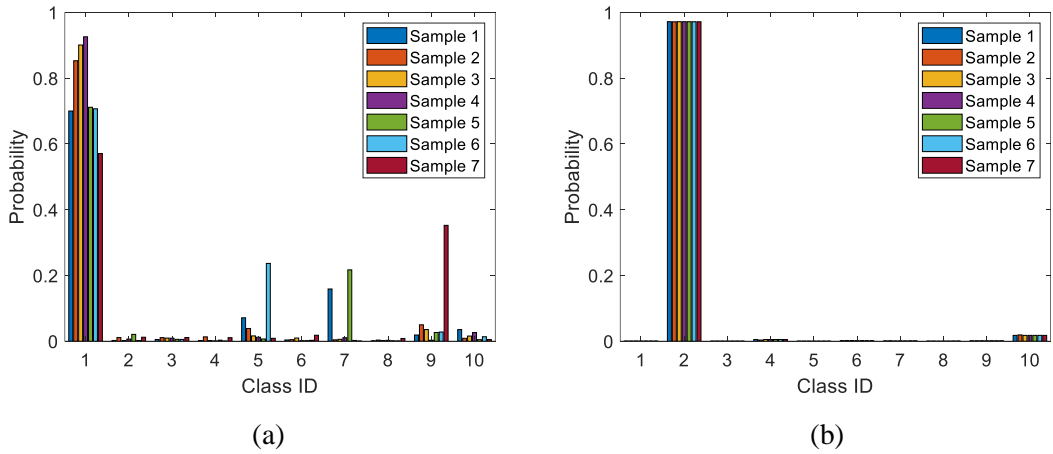

(a) (b)

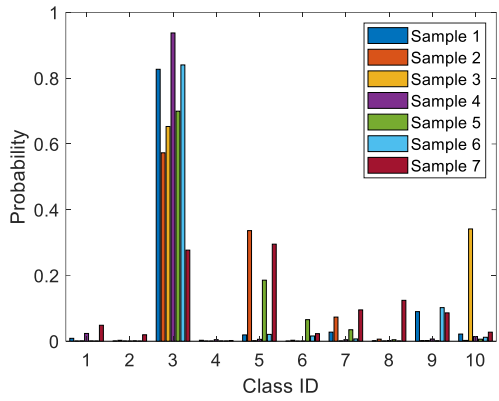
(c)

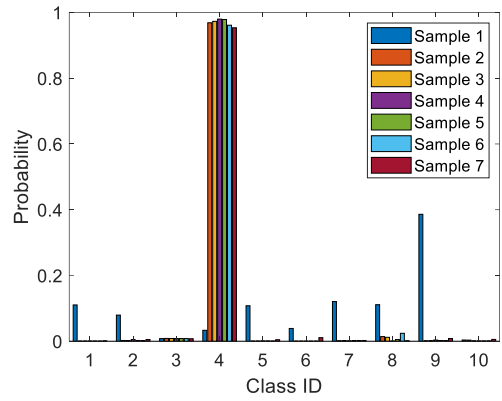
(d)

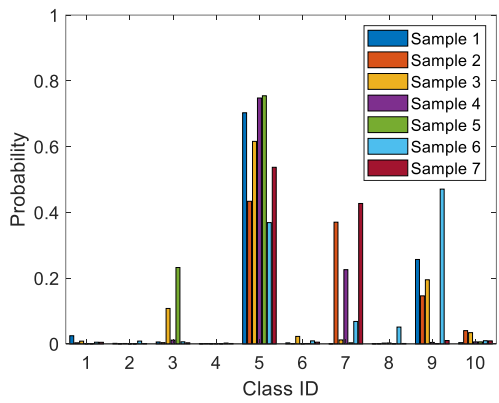
(e)

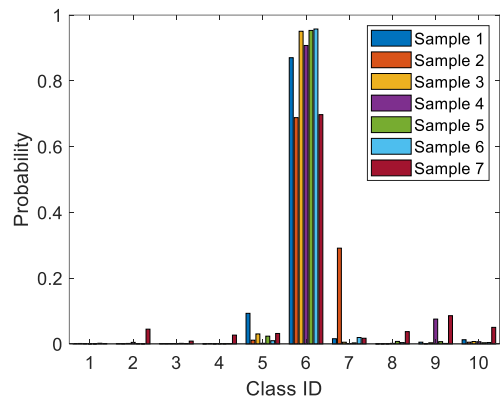
(f)

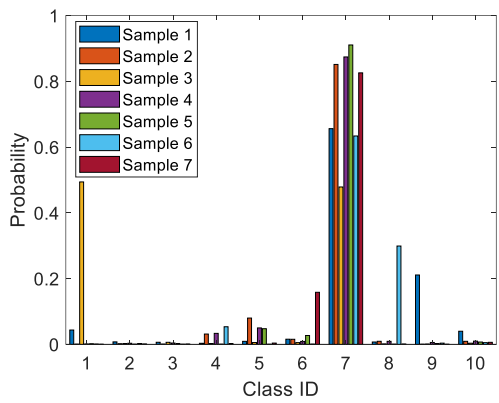
(g)

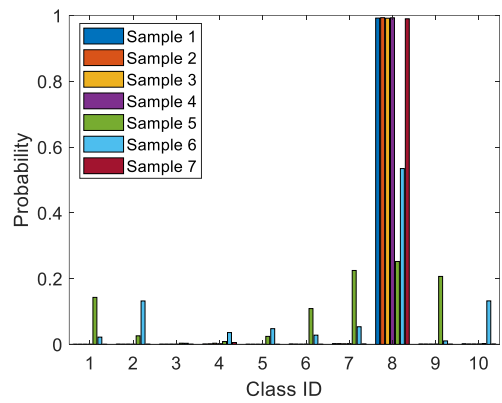
(h)

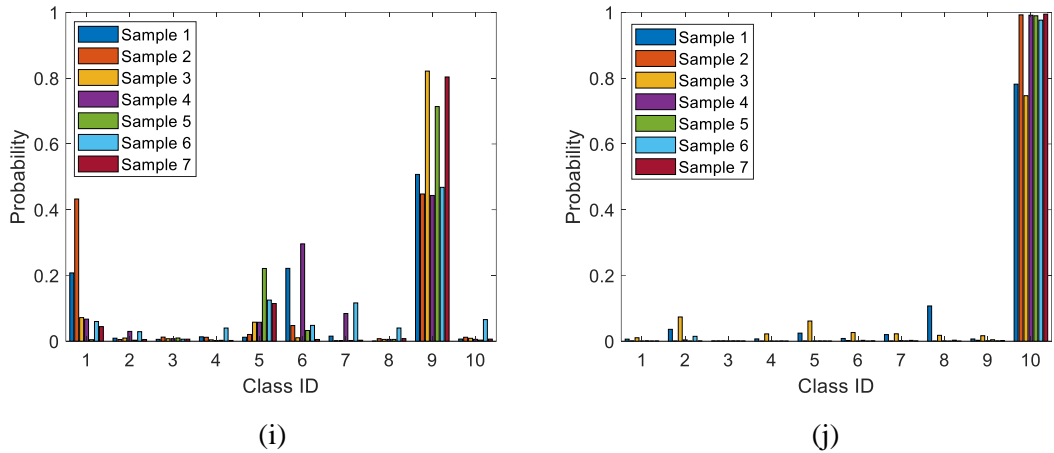

(i)                                              (j)

Figure 6. Probability of predicted classes with respect to testing samples that belong to individual class (a) Samples of class 1; (b) Samples of class 2; (c) Samples of class 3; (d) Samples of class 4; (e) Samples of class 5; (f) Samples of class 6; (g) Samples of class 7; (h) Samples of class 8; (i) Samples of class 9; (j) Samples of class 10.

While the probabilistic approach is suited to the real-world problems with probabilistic nature, a crispy metric still is required to assess the classification performance in a quantitative manner. Without loss of generality, we can simply assign the class to the testing sample when the associated probability predicted is highest, through which the above probabilistic result can be converted into the deterministic result. A confusion matrix of classification result is given in Figure 7. Obviously, the classification accuracy, i.e., the ratio of number of correct predictions to the total number of samples is 94.3%. Such classification accuracy overall is desirable. Specifically, 4 out of 70 testing samples are subject to misclassification, i.e., one sample from each of classes 3, 4, 5, 7. We revisit the probability information of misclassified samples shown in Table 2. As can be seen, except the sample 2 that is completely wrong, the rest of samples have close probability between the actual and wrongly predicted classes. This shows the possibility to further enhance the diagnosis performance if the empirical experience and judgement play a role.

|   | 1 | 2 | 3 | 4 | 5 | 6 | 7 | 8 | 9 | 10 |   |
|---|---|---|---|---|---|---|---|---|---|----|---|
| 1 | 7<br>10.0% | 0<br>0.0% | 0<br>0.0% | 0<br>0.0% | 0<br>0.0% | 0<br>0.0% | 1<br>1.4% | 0<br>0.0% | 0<br>0.0% | 0<br>0.0% | 87.5%<br>12.5% |
| 2 | 0<br>0.0% | 7<br>10.0% | 0<br>0.0% | 0<br>0.0% | 0<br>0.0% | 0<br>0.0% | 0<br>0.0% | 0<br>0.0% | 0<br>0.0% | 0<br>0.0% | 100%<br>0.0% |
| 3 | 0<br>0.0% | 0<br>0.0% | 6<br>8.6% | 0<br>0.0% | 0<br>0.0% | 0<br>0.0% | 0<br>0.0% | 0<br>0.0% | 0<br>0.0% | 0<br>0.0% | 100%<br>0.0% |
| 4 | 0<br>0.0% | 0<br>0.0% | 0<br>0.0% | 6<br>8.6% | 0<br>0.0% | 0<br>0.0% | 0<br>0.0% | 0<br>0.0% | 0<br>0.0% | 0<br>0.0% | 100%<br>0.0% |
| 5 | 0<br>0.0% | 0<br>0.0% | 1<br>1.4% | 0<br>0.0% | 6<br>8.6% | 0<br>0.0% | 0<br>0.0% | 0<br>0.0% | 0<br>0.0% | 0<br>0.0% | 85.7%<br>14.3% |
| 6 | 0<br>0.0% | 0<br>0.0% | 0<br>0.0% | 0<br>0.0% | 0<br>0.0% | 7<br>10.0% | 0<br>0.0% | 0<br>0.0% | 0<br>0.0% | 0<br>0.0% | 100%<br>0.0% |
| 7 | 0<br>0.0% | 0<br>0.0% | 0<br>0.0% | 0<br>0.0% | 0<br>0.0% | 0<br>0.0% | 6<br>8.6% | 0<br>0.0% | 0<br>0.0% | 0<br>0.0% | 100%<br>0.0% |
| 8 | 0<br>0.0% | 0<br>0.0% | 0<br>0.0% | 0<br>0.0% | 0<br>0.0% | 0<br>0.0% | 0<br>0.0% | 7<br>10.0% | 0<br>0.0% | 0<br>0.0% | 100%<br>0.0% |
| 9 | 0<br>0.0% | 0<br>0.0% | 0<br>0.0% | 1<br>1.4% | 1<br>1.4% | 0<br>0.0% | 0<br>0.0% | 0<br>0.0% | 7<br>10.0% | 0<br>0.0% | 77.8%<br>22.2% |
| 10 | 0<br>0.0% | 0<br>0.0% | 0<br>0.0% | 0<br>0.0% | 0<br>0.0% | 0<br>0.0% | 0<br>0.0% | 0<br>0.0% | 0<br>0.0% | 7<br>10.0% | 100%<br>0.0% |
|   | 100%<br>0.0% | 100%<br>0.0% | 85.7%<br>14.3% | 85.7%<br>14.3% | 85.7%<br>14.3% | 100%<br>0.0% | 85.7%<br>14.3% | 100%<br>0.0% | 100%<br>0.0% | 100%<br>0.0% | 94.3%<br>5.7% |

Output Class (y-axis) / Target Class (x-axis)

Figure 7. Confusion matrix of classifcation result.

**Table 2**. Classification probability of misclassified samples.

| Sample ID | Actual Class and Probability | | Predicted Class and Probability | |
|---|---|---|---|---|
| 1 | 3 | 0.2771 | 7 | 0.2953 |
| 2 | 4 | 0.0334 | 1 | 0.3861 |
| 3 | 5 | 0.3688 | 6 | 0.4707 |
| 4 | 7 | 0.4787 | 3 | 0.4940 |

### *3.2. Fault diagnosis performance examination and validation*

In this section, the fault diagnosis performance of the proposed methodology will be thoroughly validated. The diagnosis accuracy of GPC will be firstly compared with that of the baseline models, followed by the investigation of various influencing factors that will affect the diagnosis performance.

*3.2.1. Performance comparison with respect to the baseline meta-models under various feature extraction methods*

As mentioned previously, we let the tailored feature extraction method be adaptively selected from a specified method set in order to fully leverage the inference capability of GPC. This process is guided by the cross-validation result. In addition to analyzing the effect of feature extraction, other well-established meta-models are involved as baseline for comparison purpose. The meta-models include support vector machine (SVM) and decision tree (DT). Similar to GPC, they also need to be trained upon the limited number of features due to their algorithmic nature. Therefore, the feature extraction is necessarily applied. The cross-validation analysis that accounts for the collective influence of both feature extraction method and meta-model will be implemented. The same training and testing split ratio in Section 3.1 is adopted. The stratified random sampling cross-validation [48] with 5 emulation runs is specifically employed, and the comprehensive results are tabulated in Table 3. The mean and standard deviation of the classification accuracy among different emulations are used to evaluate the overall performance of the method combinations. As can be clearly seen, GPC + Gaussian kernel PCA is the best combination that yields the highest classification accuracy. The good performance is also robust because of very small standard deviation of accuracy. Regardless of feature extraction method, GPC significantly outperforms SVM and DT in terms of classification accuracy mean. Moreover, it is found that the feature extraction method indeed is important to the meta-model. In this case, Gaussian kernel PCA appears to be the best one among the given feature extraction methods. In comparison, stacked autoencoder is a bit inferior with respect to Gaussian kernel PCA, and the exponential kernel PCA clearly has the worst performance. The results drawn in this systematic performance examination illustrate that GPC has better classification performance than its counterparts, which readily verifies its effectiveness.

**Table 3**. Cross-validation classification accuracy of different combinaions of meta-models and feature extraction methods.

| | | Feature Extraction Method Set | | | | |
|---|---|---|---|---|---|---|
| | | Stacked Autoencoder | KPCA | | | |
| | | | Linear | Polynomial | Gaussian | Exponential |
| Meta-Model Set | GPC | Mean: **85.72** <br> 82.86, 84.29, 91.43, 84.29, 85.71 <br> Std: **3.35** | Mean: **71.43** <br> 68.57, 70.00, 70.00, 75.71, 72.86 <br> Std: **2.86** | Mean: **78.85** <br> 71.43, 72.85, 82.85, 80.00, 87.14 <br> Std: **6.65** | Mean: **92.29** <br> 94.29, 92.85, 91.43, 91.43, 91.43 <br> Std: **1.28** | Mean: **46.86** <br> 50.00, 41.43, 45.71, 52.86, 44.29 <br> Std: **4.56** |
| | SVM | Mean: **60.86** <br> 62.86, 52.86, 62.86, 62.86, 62.86 <br> Std: **4.47** | Mean: **47.14** <br> 42.86, 42.86, 48.57, 54.29, 47.14 <br> Std: **4.74** | Mean: **58.29** <br> 50.00, 60.00, 61.43, 55.71, 64.29 <br> Std: **5.57** | Mean: **83.71** <br> 75.71, 85.71, 88.57, 84.29, 84.29 <br> Std: **4.80** | Mean: **30.00** <br> 31.43, 31.43, 30.00, 30.00, 27.14 <br> Std: **1.75** |
| | DT | Mean: **70.57** <br> 70.00, 72.86, 71.43, 75.71, 62.86 <br> Std: **4.80** | Mean: **38.00** <br> 37.14, 40.00, 32.86, 45.71, 34.29 <br> Std: **5.11** | Mean: **57.14** <br> 48.57, 55.71, 57.14, 57.14, 67.14 <br> Std: **6.62** | Mean: **72.85** <br> 75.71, 70.00, 64.28, 75.71, 78.57 <br> Std: **5.72** | Mean: **31.14** <br> 40.00, 27.14, 30.00, 30.00, 28.57 <br> Std: **5.09** |

*3.2.2. Performance comparison with respect to the deep learning models*

In section 3.2.1, the advantage of GPC is reflected by comparing it with its counterparts that also learn the reduced-dimensional features. As highlighted in Introduction, deep learning neural networks recently have become the mainstream for fault diagnosis applications. As another category of meta-models, deep learning models can extract the original high-dimensional features simultaneously and automatically. Therefore, no specific feature extraction method needs to be integrated. In this research, the performance of the deep learning models also is examined and compared with that of the proposed methodology. As the neural network architecture plays an important role in the classification performance and the architecture design requires the extensive experience or trial-and-error attempts, here we directly borrow the architecture of a well-established neural network so called AlexNet [49]. The first model essentially is the transfer learning AlexNet as shown in Figure 8a. We use the transfer learning to minimize the likelihood of overfitting caused by the relatively small-sized dataset. In this transfer learning analysis, we suppose that only the parameters of last 3 full connected (FC) layers are trainable. The second model is a convolutional neural network (CNN), i.e., the former part of AlexNet as suggested by literature [50]. Because the scale of this model is much reduced as compared with that of the AlexNet, we plan to train this model from scratch. The time-series images are directly used for training. For the ease of training, the image is converted and resized as $227 \times 227$. The same amplitude scale is used for all the images, and the axis and label information in the image will be removed to avoid the learning inference. To mimic the fusion of data from different accelerometers, i.e., accelerometers at DE and FE, we combine the respective signals at the same time sequence as illustrated in Figure 9. As compared with the feature-level fusion technique adopted above, this in fact is the data-level fusion technique [47]. For healthy condition, the two-piece signals are very close (Figure 9a). They however will become notably different when the fault occurs (Figure 10a). Particularly, DE signal seems more sensitive to the outer ring (OR) fault than FE signal. Intuitively, the signal combination can increase the feature diversity and thus can improve the model performance. Such effect of data fusion will be investigated in the succeeding section.

The cross-validation analysis with the same setup shown above is carried out to yield the result in Figure 10. The scatter points denoted by the marker "red cross" represent the accuracy of different emulations, and the statistical properties, such as median, $25^{th}$ and $75^{th}$ percentile of classification accuracy can be identified upon the box information. Apparently, the CNN generally has the worst accuracy. Transfer learning AlexNet indeed leads to the performance enhancement under small-sized dataset. Overall, the performance of the proposed methodology in this research is close to that of the transfer learning AlexNet. The proposed methodology yields the narrower accuracy distribution than the transfer learning AlexNet even its best accuracy is slightly lower. While it is highly possible that the further performance enhancement of the transfer learning can be achieved by continuous tuning, such as identifying a different set of layers to be frozen during training, the whole process usually is time consuming and strongly experience-dependent. This additional set of results again illustrates the feasibility and implementation simplicity of the proposed methodology.

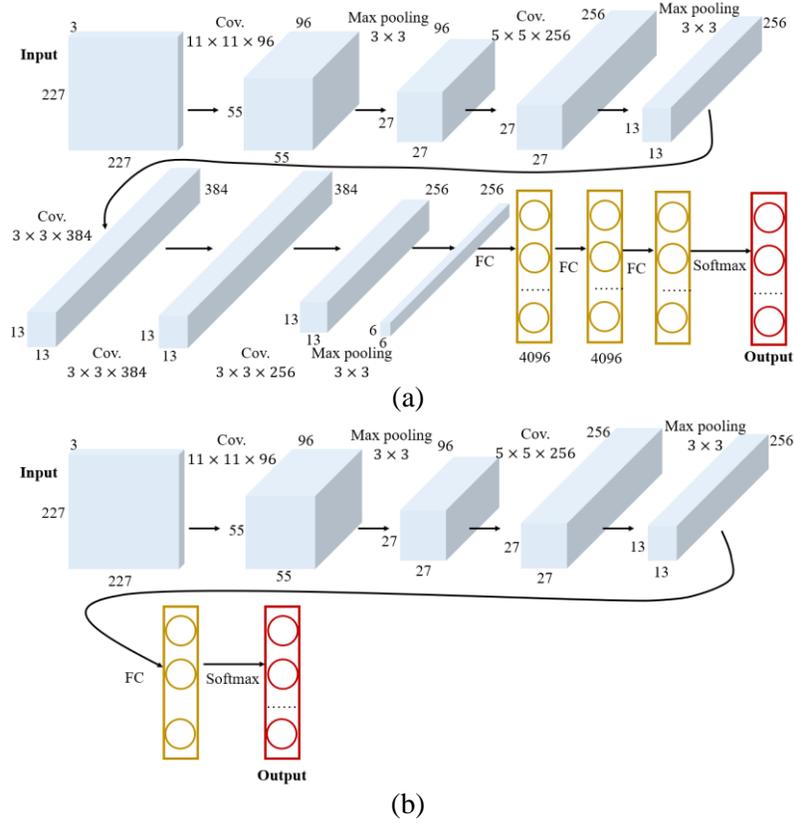

(a)

(b)

Figure 8. Architectures of proposed deep learning models (a) AlexNet transfer learning (b) CNN (local AlexNet).

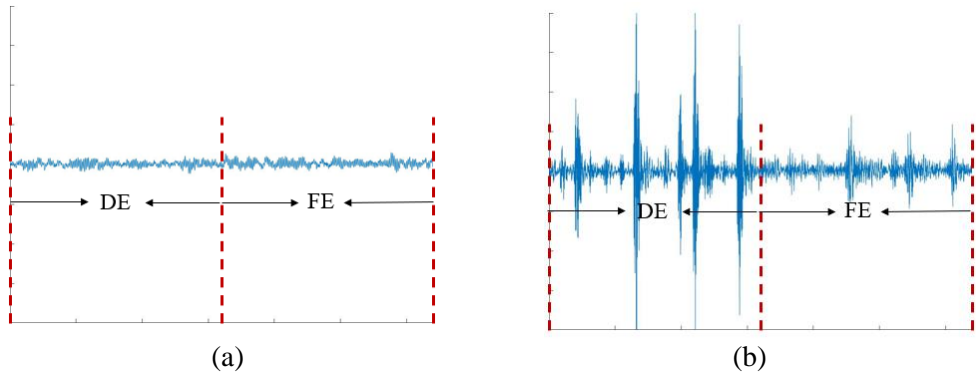

(a)                            (b)

Figure 9. Illustration of images with signal combination for deep learning analysis (a) Healthy (b) Outer ring (OR) fault (0.021 in).

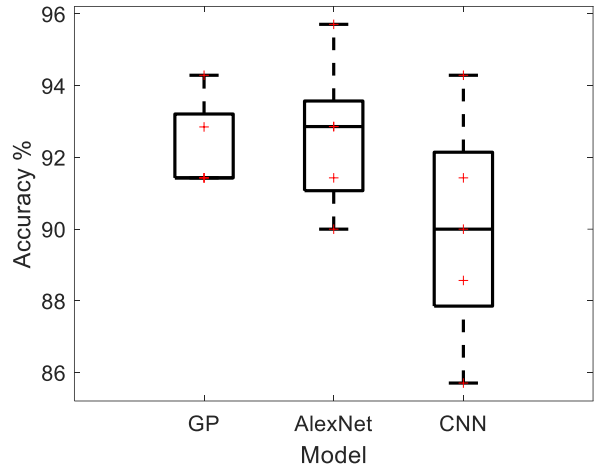

Figure 10. Cross-validation classification accuracy comparison with respect to deep learning methods.

*3.2.3. Effect of dataset configuration and quality*

As mentioned, we combine the signals of different accelerometers to increase the number of features in each data sample. We think such sensor fusion is favorable based upon the hypothesis that it is able to enrich the fault signature information ang thus improve the fault diagnosis performance. To validate this hypothesis, in this section we decide to look into another two scenarios where only the signal measured from one accelerometer either placed at DE or FE is introduced in fault diagnosis analysis. The cross-validation accuracy of the scenarios corresponding to different dataset configurations are demonstrated in Figure 11. The original result is labeled as "DE+FE", whereas the new results are labeled as "DE" and "FE", respectively. Clearly, the signal combination (DE+PE) plays a positive role in improving the classification performance. Additionally, the DE signal is more useful than FE signal in terms of accuracy. Recall Figure 1. The testing rolling bearing is installed at the DE in this research. Hence, the accelerometer at the DE can capture the vibration response that is sensitive to the fault. However, the sensitivity of the measurement from the accelerometer at the FE with respect to the fault will reduce due to the increasing wave propagation distance.

It is worth highlighting that the major strength of the proposed methodology is to conduct the probabilistic fault diagnosis by taking various uncertainty sources into account. In this research, the uncertainties include the lack of knowledge i.e., epistemic uncertainty toward the intrinsic correlation between the fault features and associated fault conditions, and the inevitable measurement error/noise i.e., aleatory uncertainty. Because the epistemic uncertainty is case dependent and oftentimes difficult to model. Here we only focus on examining the impact of the aleatory uncertainty by introducing additional white noise into the dataset. In addition to the original scenario without additional noise, two scenarios with 5% and 10 % noise are formulated. The cross-validation classification accuracy is shown in Figure 12a. The overall accuracy will decrease when the noise level increases, which agrees with the general fact that the data quality will influence the model learning process. The proposed methodology can yield around 87% accuracy mean even under very large level of noise (10%), which shows its good robustness. For each scenario, the misclassified samples can be identified for all emulation runs. The average of probability for the misclassified samples can indicate the confidence level of misclassification. Noteworthy, here the probability denotes the probability of the sample being classified to the wrong class rather than the actual/true class. The higher the probability is, the occurrence of misclassification will become more certain. The cross validation of the average of probability is given in Figure 12b. There is no clear trend of misclassification confidence level with respect to the noise level. More emulation runs may be needed to capture such trend more explicitly. This is subject to future research.

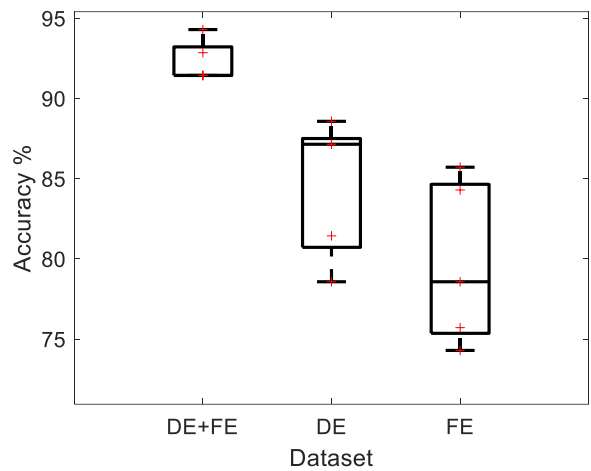

Figure 11. Cross-validation classification accuracy with respect to the dataset configuration.

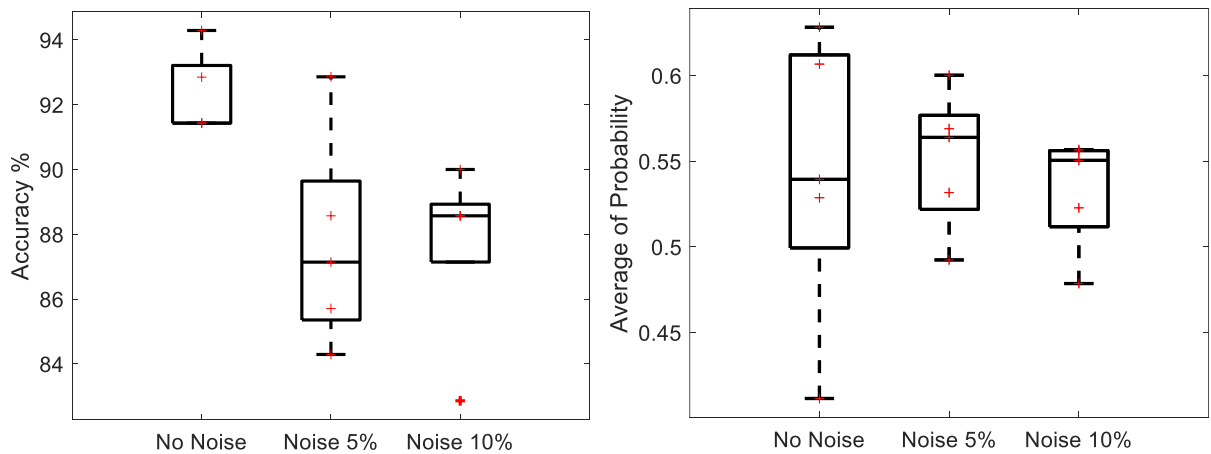

(a) (b)

Figure 12. Cross-validation classification performance with respect to the noise level introduced (a) Classification accuracy; (b) Average of probability for misclassified samples.

## 4. Conclusion

In actual bearing fault diagnosis, there exist various uncertainties that will interfere the decision making when employing the conventional deterministic classification analysis. To accommodate the practical implementation, in this research a probabilistic fault diagnosis framework with Gaussian process classifier (GPC) as the mainstay is developed. Instead of directly assigning certain known fault class to be the actual fault scenario, this framework can provide the predictive distribution of probability for each known fault class, which contains the rich probabilistic information to facilitate the subsequent decision making. Different feature extraction with dimensionality reduction methods including the KPCA methods and stacked autoencoder are incorporated into the framework, offering the flexibility to construct the high-fidelity GPC. The advantage of probabilistic classification using sensor fusion technique is clearly illustrated via the publicly accessible rolling bearing dataset, i.e., CWRU bearing dataset. The crispy classification accuracy of this framework also is thoroughly examined by comparing it with other representative meta-models. The results in case analyses validate the effectiveness of this framework.